\documentclass{article}

\usepackage{arxiv}

\usepackage[utf8]{inputenc} 
\usepackage[T1]{fontenc}    
\usepackage{hyperref}       
\usepackage{url}            
\usepackage{booktabs}       
\usepackage{amsfonts}       
\usepackage{nicefrac}       
\usepackage{microtype}      
\usepackage{lipsum}		
\usepackage{graphicx}
\usepackage{natbib}
\usepackage{doi}
\usepackage{amsmath}
\usepackage{xcolor}
\usepackage[table]{xcolor}
\usepackage{makecell}
\usepackage{tabularx}  
\usepackage{wrapfig}
\usepackage{subcaption} 
\usepackage[most]{tcolorbox}
\usepackage{listings}

\usepackage{fancyvrb}
\usepackage{fvextra}
\fvset{breaksymbol=}

\DefineVerbatimEnvironment{code}{Verbatim}{
  fontsize=\footnotesize,
  breaklines=true,
  breakanywhere=true
}
\newcommand{\highlightA}[1]{\texttt{\textbf{\textcolor{blue!60!black}{#1}}}}
\newcommand{\highlightB}[1]{\texttt{\textbf{\textcolor{green!60!black}{#1}}}}
\newcommand{\highlightC}[1]{\texttt{\textbf{\textcolor{yellow!60!black}{#1}}}}
\newcommand{\highlightD}[1]{\texttt{\textbf{\textcolor{orange!60!black}{#1}}}}
\newcommand{\infohighlight}[1]{\texttt{\textbf{\textcolor{red!60!black}{#1}}}}

\newcommand{\high}[1]{\texttt{\textbf{{#1}}}}

\tcbset{
  mycasebox/.style={
    colback=blue!5!white,          
    colbacktitle=blue!40!white,    
    colframe=blue!50!black,        
    coltitle=white,                
    fonttitle=\bfseries,           
    arc=3mm,                       
    enhanced,                      
    boxrule=0.7pt,                 
    title={#1}                     
  },
}

\title{Beneficial Reasoning Behaviors in Agentic Search and Effective Post-training to Obtain Them }

\author{
\textbf{Jiahe Jin} \quad
\textbf{Abhijay Paladugu} \quad
\textbf{Chenyan Xiong} \\
Language Technologies Institute, Carnegie Mellon University \\ 
\texttt{\{jjiahe,apaladug,cx\}@andrew.cmu.edu} \\
}

\begin{document}
\maketitle

\begin{abstract}
Agentic search requires large language models (LLMs) to perform multi-step search to solve complex information-seeking tasks, imposing unique challenges on their reasoning capabilities.
However, what constitutes effective reasoning for agentic search and how it can be learned remains unclear.
In this work, we first investigate the reasoning behaviors that enable success in agentic search. By comparing successful and failed trajectories via an LLM-based analysis pipeline, we identify four beneficial behaviors: Information Verification, Authority Evaluation, Adaptive Search, and Error Recovery. Building on this, we propose Behavior Priming, a training approach that equips agentic search models with these reasoning behaviors before reinforcement learning (RL). Specifically, it first performs supervised fine-tuning (SFT) on collected trajectories exhibiting the identified behaviors to cultivate these behaviors, and then applies standard RL to further improve task performance.
Experiments on Qwen3-1.7B and Llama3.2-3B-Instruct show that Behavior Priming yields relative improvements over direct RL by 37.2\% on three web benchmarks and 6.2\% on seven multi-hop QA benchmarks, and outperforms the SFT-then-RL baseline using outcome-correct trajectories for fine-tuning. Crucially, we show that these reasoning behaviors matter more than outcome correctness in the priming stage prior to RL. Further analysis reveals that Behavior Priming enhances exploration (pass@8) and test-time scaling (search step number), providing a robust foundation for RL. Our code are avalible at \url{https://github.com/cxcscmu/Behavior-Priming-for-Agentic-Search}.

\end{abstract}

\section{Introduction}

Agentic search~\citep{jin2025search, zheng2025deepresearcherscalingdeepresearch, li2025websailor} represents a new search paradigm in which large language models (LLMs) perform multi-step search actions to solve complex information needs. This process requires decomposing tasks, performing multi-step searches, and synthesizing search results into answers. Commercial systems such as ChatGPT's Deep Research and Google Search's AI Mode have rapidly gained adoption~\citep{zhou2024understanding, businessinsider2025apple, verge2025apple}, and the open-source community has also made notable progress, especially in applying reinforcement learning (RL) to train agentic search models~\citep{jin2025searchr1trainingllmsreason, zheng2025deepresearcherscalingdeepresearch}.


The core enabler of agentic search is the reasoning capability of LLMs. While prior works in mathematics and coding have shown that RL is highly effective at incentivizing reasoning~\citep{deepseekai2025deepseekr1incentivizingreasoningcapability, yang2025qwen3}, recent studies suggest that the success of RL depends heavily on whether the base models already exhibit specific reasoning patterns~\citep{yeo2025demystifying, gandhi2025cognitive, wang2025octothinker}, such as verification and backtracking in math. However, agentic search presents unique challenges by requiring multi-step interaction with dynamic environments. Models need to detect useful information among abundant search results, resolve contradictions between conflicting sources, and maintain focus on task goals across extended trajectories. Despite the rapid progress in applying RL to agentic search~\cite{jin2025searchr1trainingllmsreason, zheng2025deepresearcherscalingdeepresearch, song2025r1searcherincentivizingsearchcapability}, it remains uncertain which specific reasoning patterns are beneficial for this domain and how to systematically cultivate them in agentic models.

\begin{figure*}[t]
    \centering
    \includegraphics[width=\textwidth]{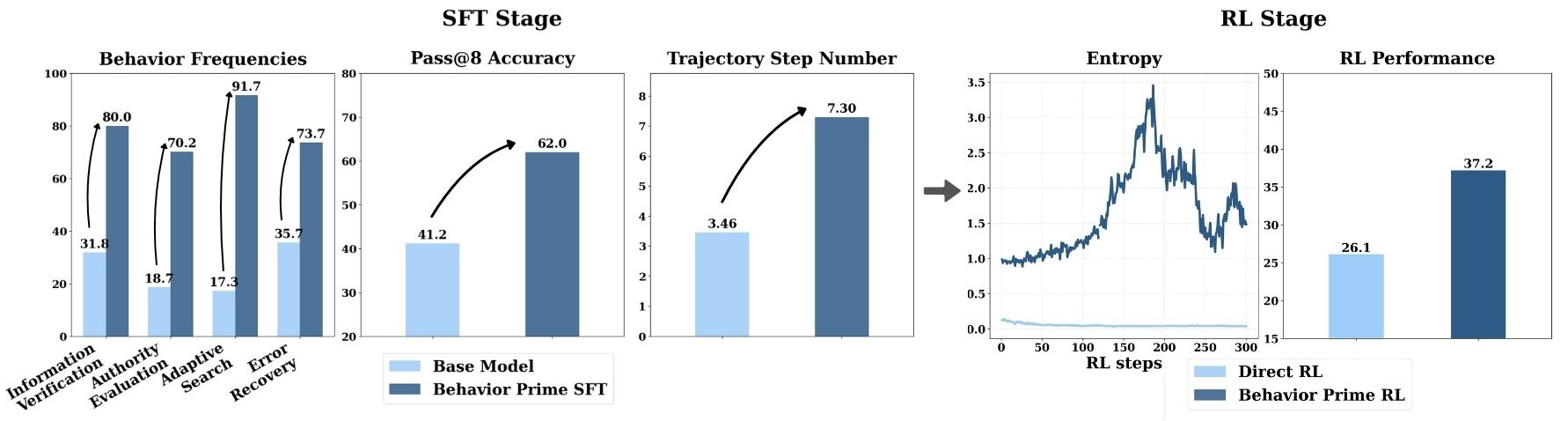}
    \caption{Training dynamics and final performance on WebWalkerQA with Qwen3-1.7B. Left (SFT Stage): The improvement of behavior frequencies, Pass@8, and average trajectory length after Behavior Priming SFT. Right (RL Stage): Entropy during RL and final RL performance for the base model (Direct RL) and the behavior-primed model.
    }
    \label{fig: overall}
\end{figure*}

In this paper, we first design an LLM-based pipeline to study effective reasoning behaviors in agentic search. We employ a reasoning LLM to compare successful and failed agentic search trajectories, and extract behaviors based on these analyses. Through this process, we identify four beneficial reasoning behaviors: \textit{Information Verification} (validating results across sources), \textit{Authority Evaluation} (assessing reliability and resolving conflicts), \textit{Adaptive Search} (modifying strategies dynamically), and \textit{Error Recovery} (detecting and correcting mistakes). Evaluations across diverse models and benchmarks show that these behaviors generalize consistently: models that exhibit them more frequently achieve stronger agentic performance, indicating that their importance is not tied to a specific model or dataset.

We then propose Behavior Priming, a two-stage training method that first cultivates these reasoning behaviors in agentic search models via SFT, and then further improves performance through RL. Specifically, we construct a large corpus of agentic search trajectories, filter for those that exhibit all four target reasoning behaviors, and fine-tune models on the resulting data.  Then we apply RL on the behavior-primed model, allowing it to refine and internalize these behaviors through its own interaction experience, ultimately translating them into improved task performance.

We demonstrate the effectiveness of Behavior Priming through experiments on Qwen3-1.7B and Llama3.2-3B-Instruct across three web benchmarks and seven multi-hop QA benchmarks. We compare our method against direct RL and two SFT-then-RL baselines, including SFT on correct-outcome trajectories and direct distillation. Results demonstrate that Behavior Priming yields relative improvements over direct RL by 37.2\% on web benchmarks and 6.2\% on multi-hop QA benchmarks, and consistently outperforms two SFT-then-RL baselines.

Crucially, we conducted an ablation study to disentangle the impact of reasoning behaviors versus outcome correctness, revealing that behavior-priming is effective even when the trajectories containing reasoning behaviors lead to incorrect outcomes. We compare models fine-tuned on trajectories that exhibit desirable behaviors but lead to opposing outcomes: correct versus incorrect answers. Remarkably, the model behavior-primed on incorrect trajectories achieves performance comparable to its counterpart behavior-primed on correct ones. This surprising result highlights that reasoning behaviors, rather than outcome correctness, are the critical factor for unlocking RL potential.


Our analysis of training dynamics further clarifies the mechanisms behind Behavior Priming. As shown in Figure~\ref{fig: overall}, after the SFT stage, the frequency of beneficial reasoning behaviors significantly increases, accompanied by a concurrent rise in pass@8 accuracy and the average number of steps per trajectory. This provides a robust foundation for exploration and test-time scaling in subsequent RL. Throughout the RL stage, the behavior-primed model maintains higher policy entropy, whereas the non-primed model with direct RL suffers from a steep decline in entropy and premature convergence. As a result, the behavior-primed model achieves substantially stronger RL performance. We also observe that the non-primed models fail to develop these reasoning behaviors endogenously during RL, highlighting the necessity of an explicit behavioral prior.

Our key contributions are as follows:
\begin{enumerate}
    \item We identify four beneficial reasoning behaviors for agentic search with an LLM-based pipeline that compares trajectories from multiple models.
    \item We propose Behavior Priming, a technique that first instills beneficial reasoning behaviors via SFT and then boosts model performance with RL.
   \item We empirically demonstrate that Behavior Priming significantly unlocks a model's potential in RL, enabling higher performance by establishing a robust foundation for exploration and test-time scaling capabilities.
\end{enumerate}

\section{Related Work}

Agentic search is an emerging search paradigm where LLMs iteratively use search-related tools to gather external information for solving complex, information-seeking tasks~\cite{xu2025comprehensive, gptsearch, claudesearch}. These systems can be categorized into two approaches. The first is multi-agent systems~\cite{alzubi2025open, GPTResearcher, li2025webthinker} with a meticulously pre-defined workflow. The second approach focuses on single-agent, end-to-end systems~\cite{zheng2025deepresearcherscalingdeepresearch, jin2025search, nguyen2025sfr} where a single LLM iteratively invokes  search-related tools based history step context. Research on agentic search training has predominantly concentrated on the latter one as it's more easy and effective for end-to-end training.

To train agentic search models, prior work relies on SFT with synthesized trajectories, where strong LLMs generate reasoning and search traces that are distilled into smaller models~\cite{asai2023selfraglearningretrievegenerate, wang2025chainofretrievalaugmentedgeneration}. More recent studies explore RL methods: some apply offline methods such as DPO in multi-agent systems~\cite{li2025webthinker}, and most approaches train single-agent models end-to-end with online RL~\cite{zheng2025deepresearcherscalingdeepresearch, jin2025searchr1trainingllmsreason}. To improve training efficiency, several works further adopt a two-stage SFT-then-RL paradigm, using SFT to initialize fundamental tool-use and reasoning capabilities before RL optimization~\cite{li2025websailor, tan2025ragr1incentivizingsearchreasoning}.

From a broader perspective on LLM reasoning, RL has been shown to substantially improve performance in domains such as mathematics and coding~\cite{deepseekai2025deepseekr1incentivizingreasoningcapability}. Recent works have shown that the effectiveness of RL is closely tied to a model’s initial reasoning properties: models that benefit significantly from RL usually already exhibit certain behaviors like verification and backtracking before RL~\cite{gandhi2025cognitive, liu2025understanding, yeo2025demystifying}. However, relevant research has been predominantly concentrated on mathematics. For agentic domains, the beneficial reasoning behaviors that enable effective RL remain underexplored. We study this question in the context of agentic search.



\section{Identifying Beneficial Behaviors in Agentic Search}

To investigate what constitutes effective reasoning in agentic search, we first develop a standard agentic search framework as a foundation for subsequent studies (Section~\ref{sec: agent_framework}), and then design a trajectory analysis pipeline to identify beneficial reasoning behaviors (Section~\ref{sec: identify_behavior}).

\subsection{Standard Agenic Search Framework}
\label{sec: agent_framework}

To enable a unified investigation across diverse LLMs, we introduce a standard agentic search framework that follows common practice~\cite{jin2025searchr1trainingllmsreason, song2025r1searcherincentivizingsearchcapability, zheng2025deepresearcherscalingdeepresearch}. The framework operates iteratively. At step $k$, given the history context $ctx_k$, the model produces an output $y_k = \langle t_k, a_k \rangle$, where $t_k$ is the reasoning process and $a_k$ is the selected action. Actions are chosen from \texttt{search} and \texttt{answer}, with an additional \texttt{summary} action for context management. Specifically, \texttt{search} retrieves external information $info_k$, \texttt{answer} terminates the interaction, and \texttt{summary} compresses the accumulated history to manage context length, enabling support for models with limited context capacity. The complete prompts are provided in Appendix~\ref{sec: agent_prompt}.

Formally, the history context is updated as:

{\small
\begin{equation*}
ctx_{k+1} =
\begin{cases}
  ctx_k + y_k + info_k, & \text{if } a_k = \texttt{search} \\
  a_k, & \text{if } a_k = \texttt{summary}
\end{cases}
\end{equation*}
}



\subsection{Identify Beneficial Reasoning Behaviors}
\label{sec: identify_behavior}


To identify reasoning behaviors associated with strong agentic search capabilities, we develop an LLM-based analysis pipeline to uncover frequently exhibited patterns that distinguish successful trajectories from failed ones.

We first collect trajectory pairs with differing capability levels by integrating Gemini 2.5 Flash (as a strong model) and Qwen3-1.7B (as a weak model) into our agentic search framework. Both models are evaluated on the web agent SFT dataset of~\cite{li2025chain}, which is the same dataset later used for trajectory corpus generation (Section~\ref{sec:sft}). We then randomly select 500 questions for which Gemini 2.5 Flash succeeds while Qwen3-1.7B fails, forming paired successful (with correct final answers) and failed (with incorrect final answers) trajectories. 


Next, we use reasoning LLMs to compare these trajectory pairs and extract common reasoning behaviors associated with successful outcomes. Inspired by prior work on automated rule generation~\citep{wang2025autorulereasoningchainofthoughtextracted}, we employ a three-stage pipeline: 
(1) Trajectory comparison: we compare behaviors in successful and failed trajectories to give a detailed analysis on why one attempt succeeds while the other fails; (2) Behavior extraction: we extract key reasoning behaviors contributing to success based on this analysis; (3) Behavior consolidation: we consolidate the extracted behaviors across all trajectories by removing duplicates, merging similar behaviors, and retaining broadly applicable ones. Prompts are provided in Appendix~\ref{sec: rule_prompt}. 
We used Gemini 2.5 Flash for the first two stages and Gemini 2.5 Pro for the final consolidation stage. A subsequent manual review confirms that the identified behaviors recur frequently across successful trajectories.

\begin{tcolorbox}[mycasebox={Examples of four behaviors \label{box:behaviors}}]
\label{tab: behavior_eg}
\textbf{Information Verification}:\\
\emph{``My task is clear: verify if the quoted text exactly matches Greetham's article… accuracy is paramount; I'll use `uncoupled', `authors', `mis-transmission', and `veil' to zero in on the relevant section.''}

\medskip
\textbf{Authority Evaluation}:\\
\emph{``I'm aiming for the USGS's own reports or databases, like the
`Nonindigenous Aquatic Species' page, to get the most reliable data.''}

\medskip
\textbf{Adaptive Search}:\\
\emph{`The search engine may not have indexed the quote perfectly, or the user's quote may differ slightly… I'll refine my strategy.''}

\medskip
\textbf{Error Recovery}:\\
\emph{``I realize I added an irrelevant keyword, "Teresa Teng" to my previous search... I should remove it as it is unrelated to my task.''}
\end{tcolorbox}

Through this process, we identify four reasoning behaviors, with examples provided above: 
\begin{enumerate}
    \item \textit{Information Verification}, validating search results across multiple sources, and citing evidence in the reasoning process;
    \item \textit{Authority Evaluation}, identifying conflicts among different search results and analyzing source credibility to prioritize authoritative information; 
    \item \textit{Adaptive Search}, dynamically modifying search strategies based on previous search results;
    \item \textit{Error Recovery}, recognizing and correcting mistakes made in prior steps. 
\end{enumerate}

\begin{figure}[h]
    \centering
    \includegraphics[width=\textwidth]{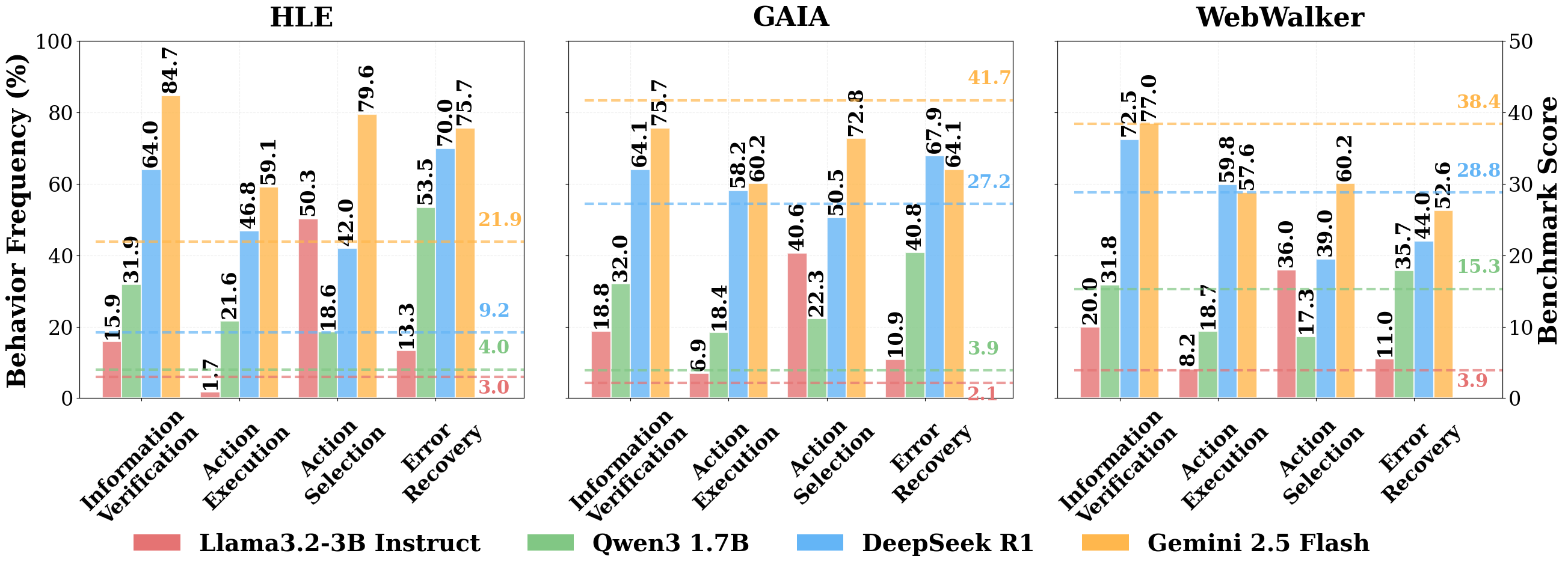}
    \caption{
    Comparison of different LLMs as the underlying agentic search model of our agent framework across three web agent benchmarks.
    For each benchmark, bars indicate the frequency of four behaviors observed in trajectories (left axis), while dashed horizontal lines indicate the benchmark score (right axis).
    }
    \label{fig:behavior_comparison}
\end{figure}

To assess the generality of these behaviors, we measure their frequency across multiple models on three downstream web agent benchmarks. Behavior frequency is defined as the proportion of trajectories that exhibit a given reasoning behavior. The prompt used for this measurement is provided in Appendix~\ref{sec: behavior_prompt}. We evaluate Gemini 2.5 Flash, DeepSeek-R1, Llama3.2-3B-Instruct, and Qwen3-1.7B. As shown in Figure~\ref{fig:behavior_comparison}, we observe a strong correlation between model performance and behavior frequency. The consistency of this trend across multiple models and heterogeneous downstream datasets provides compelling evidence that these reasoning behaviors are tightly coupled with successful agentic search performance.

\section{Behavior Priming: Unlocking the RL Potential for Agentic Search}

Having identified the beneficial reasoning behaviors, we now investigate whether we can cultivate these behaviors to improve performance. We propose ``Behavior Priming'', a method that first cultivates target behaviors and then further improves performance via RL.

\subsection{Instilling Behaviors via SFT}
\label{sec:sft}



To cultivate targeted behaviors in agentic models, we collect a trajectory corpus from a strong model, filter for trajectories that exhibit all identified reasoning behaviors, and use them for SFT. 

We first generate a trajectory corpus on two types of tasks: web agent tasks and multi-hop QA tasks. Questions and ground-truth answers for both datasets are drawn from the SFT dataset of~\cite{li2025chain}. For each question, we sample 10 trajectories using Gemini 2.5 Flash.

We then evaluate each trajectory for the presence of each behavior and select for those that exhibit all four behaviors to construct the behavior-priming dataset. For evaluation, we concatenate the model outputs and search results at each step to form a complete trajectory, and prompt an LLM to assess whether the reasoning process at any step demonstrates each target behavior. Prompt is provided in Appendix~\ref {sec: behavior_prompt}. We select trajectories that exhibit all behaviors as training data.

Following data preparation, we train models using supervised fine-tuning. We treat each step in a trajectory as an independent training sample: for a trajectory $\mathcal{T}_i = (\langle x_1^i, y_1^i \rangle, \langle x_2^i, y_2^i \rangle, \dots, \langle x_{L_i}^i, y_{L_i}^i \rangle)$ where $x_i, y_i$ are the input and output, we aggregate all input-output pairs to construct the SFT dataset:
$$
\mathcal{D}_{SFT} = \{ \langle x_k^i, y_k^i \rangle \mid 1 \le k \le L_i \}
$$

\subsection{Reinforcement Learning}
\label{rl}
Building on the cultivated reasoning behaviors, we use RL to further enhance the model's performance by allowing it to learn from its own exploration experience.


We train models with the GRPO~\citep{shao2024deepseekmath} algorithm. For each question $q$, we rollout a group of $G$ trajectories $\{\mathcal{T}_i\}_{i=1}^G$. The policy is updated by aggregating over all steps from the rollout trajectories. The loss function is defined as:

\vspace{-15pt}
\begin{equation*}
\small
\begin{aligned}
\mathcal{J}_{\mathrm{GRPO}}(\theta)
&= \mathbb{E}_{\substack{q \sim \mathcal{D},\;
\{\mathcal{T}_i\}_{i=1}^G \sim \pi_{\theta_{\text{old}}}}} \Bigg[ \sum_{i=1}^{G} \sum_{k=1}^{L_i} \sum_{t=1}^{|y_k|}\frac{1}{|y_k|}
\min \Big(
r_{i,k,t}(\theta)\hat{A}_i,\,
\text{clip}(r_{i,k,t}(\theta), 1 \pm \varepsilon)\hat{A}_i
\Big)
\Bigg]
\end{aligned}
\end{equation*}

\vspace{-10pt}

where $r_{i,k,t}(\theta)$ is the importance sampling ratio, $x_k$ and $y_k$ are the input and output at step $k$, and $\hat{A}_{i}$ is the advantage for that trajectory.

We employ an outcome-based reward. An LLM-judge evaluates the final answer $a_i$ against ground-truth, assigning a binary reward $R_i$ of 1 for a correct answer and 0 otherwise. The reward $R_i$ and the advantage $\hat{A}_i$ are constant for all steps in trajectory $\mathcal{T}_i$. For datasets, we use the web agent RL dataset of~\cite{li2025chain} for the web agent tasks and the dataset from~\cite{zheng2025deepresearcherscalingdeepresearch} for multi-hop QA tasks.

\section{Experiments}
\label{experiments}

In this section, we present a comprehensive evaluation of our Behavior Priming method. We compare it with direct RL and strong SFT-then-RL baselines on web agent and multi-hop QA tasks (Section~\ref{sec: baseline}), disentangle the reasoning process from outcome correctness (Section~\ref{sec: reasoning_vs_outcome}), study the impact of SFT data scale (Section~\ref{sec: data_size}), and analyze the necessity of composite behaviors versus single-behavior priming (Section~\ref{sec: single_behavior}). Detailed experimental setup is provided in Appendix~\ref{sec: experiment_setup}.

\subsection{Baselines}
\label{sec: baseline}
 
To evaluate the effect of our behavior-centric data curation method, we compare it against some widely adopted methods for agentic training.

\paragraph{Direct RL}
We include the \textit{Direct RL} baseline that applies RL directly without any preceding supervised fine-tuning. This baseline is introduced to explicitly assess the role of the behavior cultivation stage of SFT, and to examine whether models can spontaneously acquire the beneficial reasoning behaviors through RL, or achieve strong performance without the behavioral prior.

\paragraph{SFT-then-RL Baselines} We compare our behavior-centric method with two common SFT-then-RL baselines with different fine-tuning data selection strategies: \textit{Distillation}, which randomly samples trajectories from the corpus generated by the stronger model; and \textit{Outcome-driven}, which filters for trajectories with the correct final answer. Both baselines select trajectories for fine-tuning from the same Gemini 2.5 Flash-generated corpus as our method (Section~\ref{sec:sft}), ensuring a strong and consistent point of comparison. Detailed dataset statistics are reported in Table~\ref{tab:sft_data}.

\paragraph{Prior Works}
To evaluate our entire training framework (e.g., agent framework, datasets, RL algorithm), we compare against prior agentic search systems trained with RL, including Search-R1~\cite{jin2025searchr1trainingllmsreason}, R1-searcher~\cite{song2025r1searcherincentivizingsearchcapability}, and Deepresearcher~\cite{zheng2025deepresearcherscalingdeepresearch}. We directly adopt the evaluation results reported in~\cite{zheng2025deepresearcherscalingdeepresearch}.

\begin{table*}[t]
\centering
\small
\setlength{\tabcolsep}{3pt} 
\caption{Behavior frequencies, outcome accuracy, and trajectory statistics for the SFT datasets on both tasks. The number of steps is the number of data samples used for training (\#Steps = \#Traj. * Avg. Steps).}
\begin{tabular}{lcccc|c|ccc}
\toprule
Web Agent Task &
Info. Verif. &
Auth. Eval. &
Adapt. Search &
Error Rec. &
Accuracy &
Avg. Steps &
\#Traj. &
\#Steps \\
\midrule
Distillation & 71.7  & 42.2  & 52.3  & 36.2  & 40.0\%  & 4.6 & 4.3k & 20k\\
Outcome-driven       & 85.7 & 52.2 & 53.2 & 28.0 & 100.0\% & 3.9 & 5.1k & 20k \\
Behavior Prime           & 100.0 & 100.0 & 100.0 & 100.0  & 49.8\%  & 6.8 & 2.9k & 20k\\
Behavior Prime (Incorrect) & 100.0  & 100.0  & 100.0  & 100.0  & 0.0\%  & 7.6 & 2.6k & 20k\\
Behavior Prime (Correct) & 100.0  & 100.0  & 100.0  & 100.0  & 100.0\%  & 5.9 & 3.4k & 20k\\
\midrule
Multi-hop QA Task & & & & & & & & \\
\midrule
Distillation  & 93.0 & 29.8 & 25.6 & 4.3 & 58.1\% & 2.5 & 4.1k & 10k \\
Outcome-driven & 94.9 & 25.8 & 25.3 & 2.6 & 100.0\% & 2.5 & 4.1k & 10k \\
Behavior Prime & 100.0 & 100.0 & 100.0 & 100.0 & 61.8\% & 4.6 & 2.2k & 10k \\
\bottomrule
\end{tabular}
\label{tab:sft_data}
\end{table*}

\subsection{Disentangling Reasoning Behaviors from Outcome Correctness}
\label{sec: reasoning_vs_outcome}

To disentangle the influence of reasoning processes from final outcomes, and to assess the synergy between high-quality reasoning and outcome correctness, we evaluate two variants of our method: \textit{Behavior Prime (Incorrect)}, which fine-tunes models on trajectories exhibiting all four reasoning behaviors but resulting in incorrect outcomes; and \textit{Behavior Prime (Correct)}, which fine-tunes models on trajectories exhibiting all four behaviors and resulting in correct outcomes. Both methods apply standard RL after the SFT stage.

\subsection{Impact of the Behavior-Trajectory Data Size in SFT}
\label{sec: data_size}

We investigated how the scale of the behavior-trajectory data used in SFT affects the efficacy of Behavior Priming. We fine-tuned Qwen3-1.7B on subsets of our Behavior Prime dataset on the web task with varying sizes: 5k, 10k, and the full 20k samples. We then compared their performance against the Direct RL baseline (denoted as 0k). 

\subsection{Contrasting Single-Behavior Priming with Composite-Behavior Priming}
\label{sec: single_behavior}

We investigated whether the composite set of four behaviors was truly necessary, or if the benefit could be achieved by priming with a single behavior. We identified Information Verification (IV) as the most common behavior in our trajectory corpus and selected it for this contrastive study. We used two SFT datasets on the web task of equal size (10k samples each): 1) \textit{Behavior Prime (IV-Only)-10k}: Trajectories selected only for exhibiting Information Verification. 2) \textit{Behavior Prime-10k}: The 10k subset dataset in Section~\ref{sec: data_size} that contains all four behaviors. We then applied standard RL on both SFT checkpoints.

\section{Results}

\begin{table*}[h!]
\centering
\caption{Performance on web agent benchmarks before and after RL. The overall score is calculated as the average of seven benchmarks. The scores in the \textit{Before RL} section are performance right after the SFT stage, and the final performance scores are under in \textit{After RL} section.}
\label{tab:web_result}
\resizebox{\textwidth}{!}{%
\begin{tabular}{l l cccc | c | c | c}
\hline
& & \multicolumn{4}{c}{\textbf{GAIA}} & \multicolumn{1}{c}{\textbf{WebWalkerQA}} & \multicolumn{1}{c}{\textbf{HLE}} & \multicolumn{1}{c}{\textbf{Overall}} \\
\cline{3-6} \cline{7-7} \cline{8-8} \cline{9-9}
\textbf{Method} & \textbf{Base Model} & \textbf{Level 1} & \textbf{Level 2} & \textbf{Level 3} &
\textbf{Avg.} &
 &
 &
 \\
\hline
\rowcolor{blue!15}
\multicolumn{9}{l}{\textit{Before RL (After SFT)}} \\
Direct RL (No SFT) & Qwen3-1.7B & 7.7 & 1.9 & 0.0 & 3.9 & 15.3 & 4.0 & 7.7 \\
Distillation & Qwen3-1.7B & \textbf{12.8} & 7.7 & 0.0 & 8.7 & 23.5 & \textbf{5.4} & 12.5 \\
Outcome-driven & Qwen3-1.7B & 10.3 & \textbf{9.6} & 0.0 & 8.7 & \textbf{24.3} & 4.8 & \textbf{12.6} \\
Behavior Prime & Qwen3-1.7B & \textbf{12.8} & 7.7 & 0.0 & 8.7 & 22.0 & 4.6 & 11.8 \\
\hline
\rowcolor{blue!15}
\multicolumn{9}{l}{\textit{After RL}} \\
Direct RL & Qwen3-1.7B & 15.4 & 11.5 & 0.0 & 11.7 & 26.1 & 3.9 & 13.9 \\
Distillation  & Qwen3-1.7B & 18.0 & 11.5 & \textbf{16.7} & 14.6 & 33.2 & 7.4 & 18.4\\
Outcome-driven  & Qwen3-1.7B & 23.1 & 17.3 & 0.0 & 17.5 & 36.8 & 5.8 & 20.0 \\
\textbf{Behavior Prime} & Qwen3-1.7B & \textbf{28.2} & \textbf{21.2} & 0.0 & \textbf{21.4} & \textbf{37.2} & \textbf{7.8} & \textbf{22.3} \\
\hline
\vspace{1pt} \\[-1em] 
Direct RL & Llama3.2-3B-Instruct & 12.8 & 11.5 & 8.3 & 11.7 & 24.7 & 7.0 & 14.5 \\
Outcome-driven & Llama3.2-3B-Instruct\ & 15.4 & 11.5 & 16.7 & 13.6 & 28.8 & 9.8 & 13.3 \\
Distillation & Llama3.2-3B-Instruct & 15.4 & \textbf{21.5} & \textbf{16.7} & \textbf{18.4} & 26.5 & 6.6 & 17.2 \\
\textbf{Behavior Prime} & Llama3.2-3B-Instruct & \textbf{25.6}& 13.4 & \textbf{16.7} & \textbf{18.4} & \textbf{33.8} & \textbf{7.5} & \textbf{19.9} \\
\hline
\end{tabular}%
}
\end{table*}
\vspace{-10pt}

\begin{table*}[h!]
\centering
\caption{Performance on multi-hop QA benchmarks. The overall score is the average score of seven benchmarks.}
\label{tab:multihop_result}
\resizebox{\textwidth}{!}{%
\begin{tabular}{l l ccccccc | c}
\hline
\textbf{Method} & \textbf{Base Model} & \textbf{2wiki} & \textbf{Bamboogle} & \textbf{Hotpotqa} & \textbf{Musique} & \textbf{NQ} & \textbf{PopQA} & \textbf{TriviaQA} & \textbf{Overall} \\
\hline
\rowcolor{blue!15}
\multicolumn{10}{l}{\textit{Prior work baselines}} \\
Search-R1-base & Qwen2.5-7B-Base & 47.9 & 57.6 & 63.0 & 27.5 & 60.0 & 47.0 & 76.2 & 54.2 \\
Search-R1-instruct & Qwen2.5-7B-Instruct & 48.8 & 47.2 & 52.5 & 28.3 & 49.6 & 44.5 & 49.2 & 45.7 \\
R1-Searcher & Qwen2.5-7B-Base & 65.8 & 65.6 & 53.1 & 25.6 & 52.3 & 43.4 & 79.1 & 55.7 \\
DeepResearcher & Qwen2.5-7B-Instruct & 66.6 & 72.8 & 64.3 & 29.3 & 61.9 & 52.7 & 85.0 & 61.8 \\
\hline
\rowcolor{blue!15}
\multicolumn{10}{l}{\textit{Our Work}} \\
Direct RL & Qwen3-1.7B & 66.8 & 64.0 & 61.7 & 25.0 & 72.7 & 53.5 & \textbf{87.9} & 61.7 \\
Distillation & Qwen3-1.7B & 65.6 & \textbf{74.4} & 64.8 & 26.9 & \textbf{74.0} & 56.1 & 85.7 & 63.9 \\
Outcome-driven & Qwen3-1.7B & 70.3 & 68.8 & 65.0 & 29.7 & 72.7 & 54.7 & 86.3 & 63.9 \\
\textbf{Behavior Prime} & Qwen3-1.7B & \textbf{73.8} & 70.4 & \textbf{67.0} & \textbf{30.7} & 73.6 & \textbf{56.6} & 86.9 & \textbf{65.5} \\
\hline
\vspace{1pt} \\[-1em] 
Direct RL (No SFT) & Llama3.2-3B-Instruct & 74.0 & 72.0 & 65.8 & 31.5 & 74.8 & 56.4 & 90.4 & 66.4 \\
Distillation & Llama3.2-3B-Instruct & 77.7 & 76.0 & \textbf{71.3} & 32.0 & 78.9 & 55.7 & 89.3 & 68.7 \\
Outcome-driven & Llama3.2-3B-Instruct & 69.9 & 72.0 & 69.9 & 30.5 & 77.3 & 60.0 & 89.5 & 67.0 \\
\textbf{Behavior Prime} & Llama3.2-3B-Instruct & 75.6 & \textbf{80.0} & 69.5 & \textbf{38.1} & \textbf{82.0} & \textbf{60.4} & \textbf{90.6} & \textbf{70.9} \\
\hline
\end{tabular}%
}
\end{table*}

\begin{table*}[h!]
\centering
\caption{Performance comparison between models behavior-primed on correct vs. incorrect trajectories before RL.}
\label{tab:behavior_correct_ablation}
\resizebox{\textwidth}{!}{%
\begin{tabular}{l l cccc | c | c | c}
\hline
& & \multicolumn{4}{c}{\textbf{GAIA}} & \multicolumn{1}{c}{\textbf{WebWalkerQA}} & \multicolumn{1}{c}{\textbf{HLE}} & \multicolumn{1}{c}{\textbf{Overall}} \\
\cline{3-6} \cline{7-7} \cline{8-8} \cline{9-9}
\textbf{Method} & \textbf{Base Model} & \textbf{Level 1} & \textbf{Level 2} & \textbf{Level 3} &
\textbf{Avg.} &
 &
 &
 \\
\hline
\rowcolor{blue!15}
\multicolumn{9}{l}{\textit{Before RL (After SFT)}} \\
Behavior Prime (Incorrect) & Qwen3-1.7B & 7.7 & \textbf{11.5} & 0.0 & 8.7 & 15.3 & 3.5 & 9.3 \\
Behavior Prime (Correct) & Qwen3-1.7B & 10.3 & 9.6 & 0.0 & 8.7 & 19.1 & \textbf{6.2} & 11.3 \\
\hline
\rowcolor{blue!15}
\multicolumn{9}{l}{\textit{After RL}} \\
Behavior Prime (Incorrect) & Qwen3-1.7B & 30.8 & \textbf{26.9} & 8.3 & \textbf{27.2} & 35.5 & \textbf{7.8} & 23.5 \\
Behavior Prime (Correct) & Qwen3-1.7B & \textbf{30.8} & 23.1 & \textbf{16.7} & 25.2 & \textbf{37.8} & \textbf{7.8} & \textbf{23.6} \\
\hline
\end{tabular}%
}
\end{table*}


\subsection{Main Results}

In this section, we present our main experimental results, demonstrating that Behavior Priming consistently leads to best performance, and its effect holds even when priming trajectories yield incorrect outcomes. These results provide strong evidence that the identified reasoning behaviors are able to enable strong agentic search performance.

\paragraph{Behavior Priming Effectively Increases Headroom for RL}
As shown in Table~\ref{tab:web_result} and Table~\ref{tab:multihop_result}, Behavior Priming consistently achieves substantially higher performance than Direct RL. In particular, the overall scores on both Llama-3.2-Instruct and Qwen3-1.7B exceed those of Direct RL by over 37\% on web benchmarks and 6.2\% on multi-hop QA benchmarks. Moreover, Behavior Priming consistently outperforms strong SFT-then-RL baselines, including \textit{Distillation} and \textit{Outcome-driven}. These results demonstrate that Behavior Priming establishes a robust foundation through behavior-centric SFT, enabling more effective improvement during subsequent RL training.

In addition, comparisons with prior agentic search systems show that our training framework achieves competitive overall performance, indicating it provide a solid and reliable foundation for agentic search training.


\paragraph{Reasoning Behavior is More Important than Outcome Correctness}
As shown in Table~\ref{tab:web_result}, while \textit{Outcome-driven} achieves the strongest performance after SFT, this advantage does not lead to the best performance after RL. During the subsequent RL training, it is surpassed by our \textit{Behavior Prime} method. This demonstrated that although training with answer-correct trajectories provides a powerful boost after SFT, cultivating reasoning behaviors is the key to enabling effective exploration and achieving superior performance during the subsequent RL phase.


Notably, as shown in Table~\ref{tab:behavior_correct_ablation}, our reasoning behavior and outcome correctness disentanglement study demonstrates that, although the \textit{Behavior Prime (Incorrect)} variant achieved suboptimal performance after SFT, it ultimately achieved comparable performance with the \textit{Behavior Prime (Correct)} variant after RL. This confirms that the gains from behavior priming are driven by the reasoning patterns, independent of outcome correctness.


\paragraph{Full Behavior Set Leads to Greater Performance Gains}

\begin{table*}[ht]
\centering
\caption{Performance after RL on web agent benchmarks for priming with single behavior and all behaviors. The Behavior Prime (IV-Only)-10k is primed on the Information Verification behavior only with 10k data samples.}
\label{tab:single_behavior}
\footnotesize 
\begin{tabular}{l l c | c | c | c}
\hline
\textbf{Method} & \textbf{Base Model} & \textbf{GAIA} & \textbf{WebWalkerQA} & \textbf{HLE} & \textbf{Overall} \\
\hline
\vspace{1pt} \\[-1em]
Direct RL & Qwen3-1.7B & 11.7 & 26.1 & 3.9 & 13.9 \\
Behavior Prime (IV-Only)-10k & Qwen3-1.7B & 14.7 & 30.7 & 6.8 & 17.4 \\
Behavior Prime-10k & Qwen3-1.7B & \textbf{15.5} & \textbf{36.5} & \textbf{7.2} & \textbf{19.7} \\
\hline
\end{tabular}
\end{table*}

\begin{wrapfigure}{r}{0.4\columnwidth}
    \includegraphics[width=0.4\columnwidth]{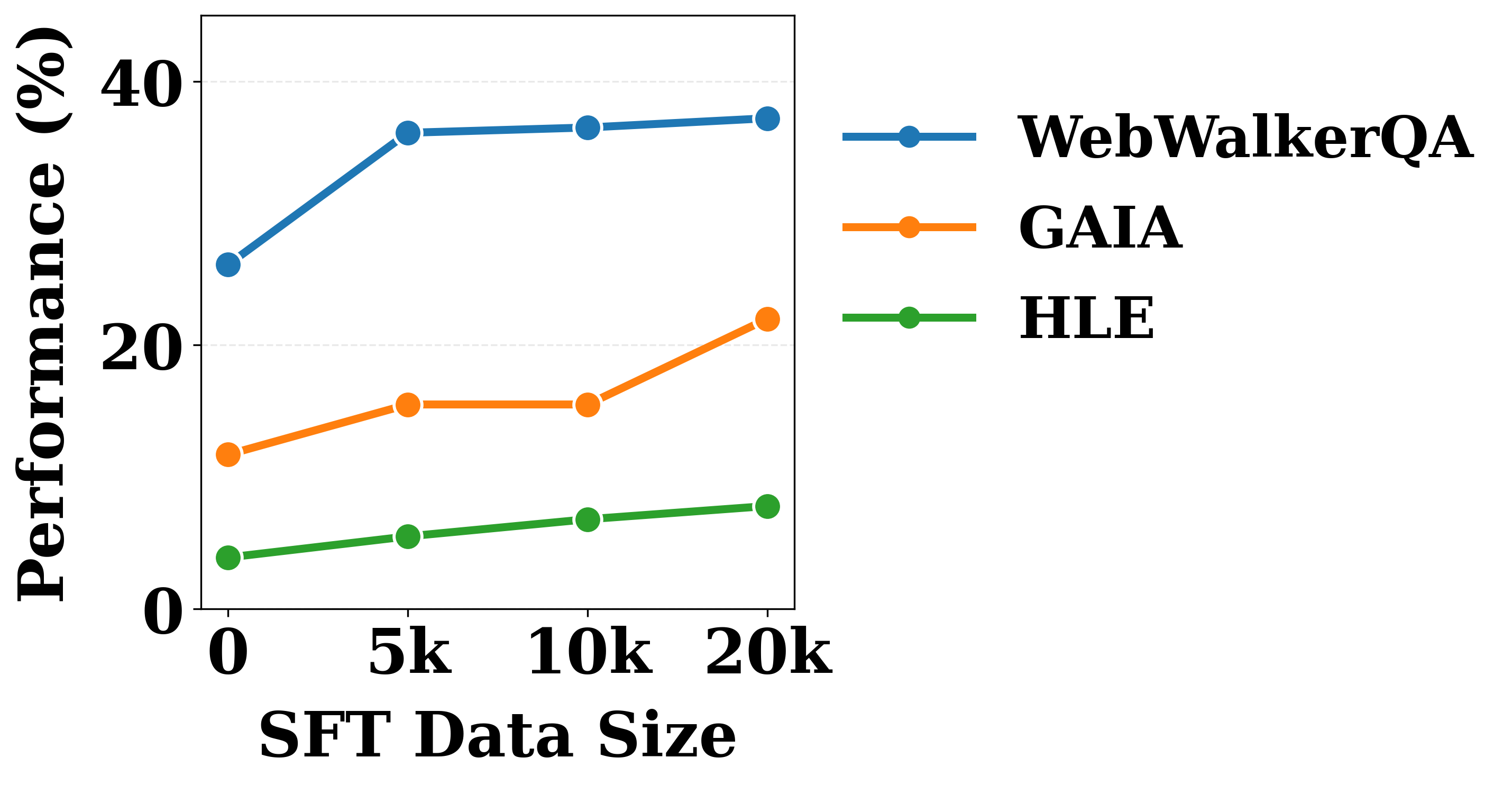}
    \caption{Qwen3-1.7B's final performance after RL when behavior primed with different data sizes in the SFT stage.}
    \label{fig:scaling_data}
\end{wrapfigure}
Results in Table~\ref{tab:single_behavior} show that while priming with only the most frequent behavior (\textit{Behavior Prime (IV-Only)-10k}) enhances performance compared to the direct RL baseline, it is consistently outperformed priming with all four behaviors (\textit{Behavior Prime-10k}). This confirms that the superior performance of our method is unlocked by instilling the complete, synergistic set of all behaviors, which are critical for building a robust foundation for RL.
\paragraph{Scaling SFT Data Enhances Behavior Priming}
Figure~\ref{fig:scaling_data} shows that increasing SFT dataset size for behavior priming improves final RL performance, indicating that the benefits of our method scale with additional data. However, performance on WebWalkerQA plateaus beyond 5k data scale. We hypothesize that the SFT stage mainly establishes a behavioral foundation for RL, and once the targeted behaviors are learned, additional priming yields diminishing returns.

\subsection{Analysis with Training Dynamics}

 \begin{figure}[h!]
    \centering
    \includegraphics[width=\textwidth]{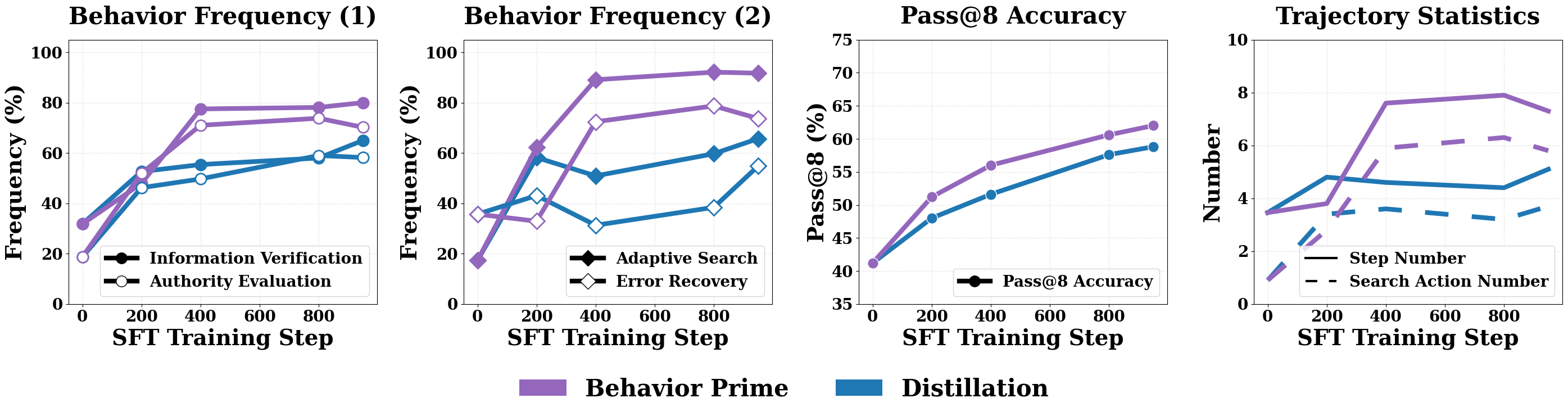}
    \caption{
    Qwen3 1.7B + \textit{Distillation} and Qwen3 1.7B + \textit{Behavior Prime}'s behavior frequencies, pass@8, and trajectories statistics (average step number and search action number per trajectory) on WebWalkerQA during the SFT process.
    }
    \label{fig:sft_trends}
\end{figure}

\begin{figure*}[h!]
    \centering
    \includegraphics[width=0.9\columnwidth]{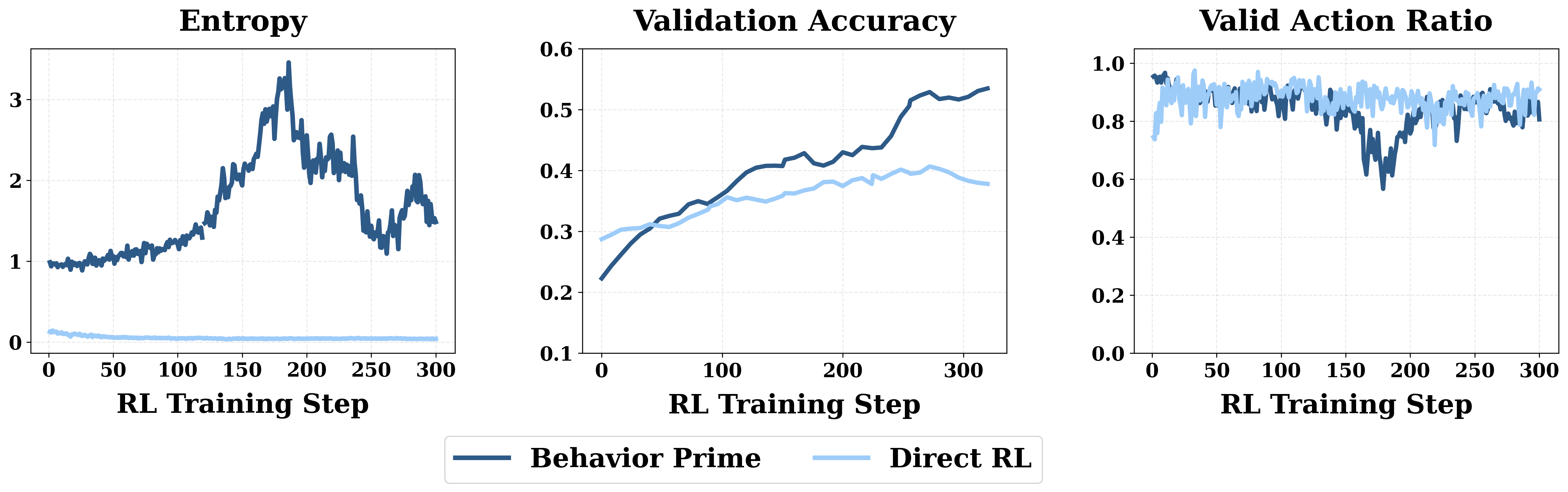}
    \caption{The entropy, validation accuracy, and valid action (percentage of  correct-format actions) during RL of Qwen3-1.7B with \textit{Behavior Prime} and \textit{Direct RL}.}
    \label{fig:rl_process}
\end{figure*}

To better understand our method, we examine how reasoning behavior frequencies, exploration abilities, trajectory length, and model performance evolve over the SFT and the RL phase.

\paragraph{Behavior Priming Encourages Exploration}
We track model evolution during SFT by monitoring behavior frequencies, Pass@8, and trajectory statistics. As shown in Figure~\ref{fig:sft_trends}, although both \textit{Distillation} and \textit{Behavior Prime} improve Pass@8 and trajectory step number alongside rising behavior frequencies, the gains are substantially larger for \textit{Behavior Priming}. This indicates that explicitly priming reasoning behaviors enables the model to explore more diverse paths and allocate greater test-time compute through more search actions.

\paragraph{Sustained Exploration Translates to a Higher RL Ceiling}

We then analyzed the training dynamics during RL, monitoring the policy entropy and validation performance. Figure~\ref{fig:rl_process} (a) shows that the behavior-primed model starts with higher entropy and is less prone to collapse. In contrast, the non-behavior-primed model (Direct RL)'s entropy collapses rapidly. 

This indicates that behavior priming endows the model with richer exploration, allowing it to seek diverse strategies rather than prematurely converging to a suboptimal policy. This sustained exploration directly translates into performance gains, as shown in Figure~\ref{fig:rl_process} (b): while the base model quickly reaches a low plateau, the behavior-primed model converges more slowly to a higher performance ceiling.

Another finding is that while the non-behavior-primed model (Direct RL) starts with a lower valid action ratio (the percentage of steps with correct action format) compared to the behavior-primed model, it successfully masters the required output format within 20 steps and maintains a high valid action ratio thereafter. Conversely, the behavior-primed model even exhibits a less stable valid action ratio during RL. This indicates that the improvements from Behavior Priming arise from enhanced reasoning behaviors rather than mere familiarity with tool-use syntax.

\section{Discussion: Is Process Rewards an Alternative for Behavior Guidance?}

\definecolor{dullgreen}{RGB}{80,200,100}

\begin{table*}[h!]
\centering
\caption{Qwen3-1.7B's behavior frequency (averages across three benchmarks) and benchmark performance with standard Direct RL and behavior-guided RL (with additional process reward encourages four reasoning behaviors).}
\label{tab:process_reward}

\setlength{\tabcolsep}{4pt}
\renewcommand{\arraystretch}{0.9}

\resizebox{\textwidth}{!}{
\begin{tabular}{llllllll}
\toprule
& \multicolumn{4}{c}{\textbf{Behavior Frequency}} & \multicolumn{3}{c}{\textbf{Benchmark Performance}} \\
\cmidrule(lr){2-5} \cmidrule(lr){6-8}
\textbf{Model} & Information & Authority & Adaptive & Error & WebWalkerQA & GAIA & HLE \\
& Verification & Evaluation & Search & Recovery \\
\midrule
Qwen3-1.7B & 28.3 & 18.1 & 18.2 & 39.7 & 15.3 & 3.9 & 4.0 \\
\midrule
Qwen3-1.7B + RL & 25.2 \textcolor{red}{(-3.1)} & 9.8 \textcolor{red}{(-8.3)} & 21.5 \textcolor{dullgreen}{(+3.3)} & 11.2 \textcolor{red}{(-28.5)} & \textbf{26.1} \textcolor{dullgreen}{(+10.8)} & \textbf{16.3} \textcolor{dullgreen}{(+12.4)} & 3.9 \textcolor{red}{(-0.1)} \\
Qwen3-1.7B + behavior guided RL & 67.3 \textcolor{dullgreen}{(+39.0)} & 50.8 \textcolor{dullgreen}{(+32.7)} & 82.8 \textcolor{dullgreen}{(+64.6)} & 86.5 \textcolor{dullgreen}{(+46.8)} & 15.4 \textcolor{dullgreen}{(+0.1)} & 7.8 \textcolor{dullgreen}{(+3.9)} & \textbf{4.4} \textcolor{dullgreen}{(+0.4)} \\
\bottomrule
\end{tabular}}
\end{table*}

We also explore alternative ways to use these beneficial behaviors to improve model performance. Curious about whether the model could spontaneously develop these reasoning behaviors during RL, we measured the change in behavior frequencies for a non-behavior-primed model after Direct RL. Results in Table~\ref{tab:process_reward} show that most of the frequencies even dropped after RL. Considering that our behavior frequency evaluation is based on process-level, we investigate whether we could guide the emergence of these behaviors with an additional process reward. Specifically, we combined the standard outcome-based reward with a process-based reward that encourages the exhibition of reasoning behaviors. We measure the number of different behaviors in a trajectory $N$, and define the final reward as $R = R_{\text{outcome}} + 0.1 \times N$.

As shown in Table~\ref{tab:process_reward}, this approach successfully increases the behavior frequency, but leads to worse performance compared to standard RL. This result aligns with previous work on math and code reasoning~\cite{deepseekai2025deepseekr1incentivizingreasoningcapability}, suggesting that the model learns to "reward hack": it mimics the surface-level patterns of the behaviors to maximize the reward but fails to grasp their essence for problem-solving. This finding reveals that our Behavior Priming method is a more effective way to improve model performance.

\section{Conclusion}
We identify four beneficial reasoning behaviors for agentic search and propose Behavior Priming, a method that cultivates these behaviors through SFT to enable more effective RL. Experiments across multiple models and ten benchmarks show that Behavior Priming achieves the strongest performance compared to other SFT-then-RL baselines, and that cultivating reasoning behaviors is more important than teaching correct answers in preparing models for RL. Further analysis reveals that Behavior Priming establishes a robust foundation for RL by enabling richer exploration, as reflected by higher Pass@8 and more search steps that support test-time scaling. Together, these findings highlight a promising direction for targeted post-training strategies that enhance reasoning behaviors and improve the effectiveness of RL in training capable agentic models.

\bibliographystyle{unsrtnat}
\bibliography{references} 
\appendix
\clearpage

\section{Detailed Experiments Setup}
\label{sec: experiment_setup}

\paragraph{Benchmarks}
For web benchmarks, we use \textbf{WebWalkerQA}~\cite{wu2025webwalkerbenchmarkingllmsweb}, \textbf{GAIA}~\cite{mialon2023gaia}(following ~\cite{li2025webthinker, wu2025webdancer, li2025chain} to use the 103 text-based examples), and \textbf{Humanity's Exam (HLE)}~\cite{phan2025humanitysexam}. For multi-hop QA benchmarks, we use \textbf{NQ}~\cite{kwiatkowski2019natural}, \textbf{TQ}~\cite{joshi2017triviaqa}, \textbf{HotpotQA}~\cite{yang2018hotpotqa}, \textbf{2Wiki}~\cite{ho2020constructing}, \textbf{MuSiQue}~\cite{trivedi2022musique}, \textbf{Bamboogle}~\cite{press2023measuring}, and \textbf{PopQA}~\cite{mallen2023trustlanguagemodelsinvestigating}. We followed~\cite{zheng2025deepresearcherscalingdeepresearch} to use 512 examples from the development sets of NQ, TQ, HotpotQA, 2Wiki, MuSiQue, and PopQA and all 125 samples from Bamboogle. 

\paragraph{Evaluation}
The max step number for the trajectory is 25 for web benchmarks and 15 for multi-hop QA benchmarks. The temperature is 0.0 for the main evaluation, and 1.0 for the pass@k evaluation. We use GPT-4o-mini to score the final answers. Prompts are provided in Appendix~\ref{sec: prompt_for_eval}.

\paragraph{Training Details}
We adopt Qwen3-1.7B~\cite{yang2025qwen3} and Llama3.2-3B-Instruct~\cite{llama3} as base models. For SFT, each model was fine-tuned for 3 epochs with a batch size of 8. The RL training was conducted on the verl-agent~\cite{feng2025groupingrouppolicyoptimizationllm} framework for 300 steps, with a batch size of 32 and a GRPO group size of 8. All RL experiments were performed on 8 NVIDIA H100 GPUs, with each training run taking approximately 20 hours to complete.

\section{Prompts}
\label{sec: prompt}
\subsection{Prompt For Agent Framework}
\label{sec: agent_prompt}

We support the integration of both models with internal thinking(like Gemini 2.5 series, Qwen3 series, DeepSeek R1) and models without internal thinking. For models without internal thinking, we explicitly prompt them first to generate a thinking process and then the action. The prompts for both models as shown below:

\begin{tcolorbox}[title=Agent Framework Prompt for Models with Internal Thinking,
                  colback=gray!5, colframe=gray!50,
                  breakable, enhanced]
\begin{code}
Your are a research assistant with the ability to perform web searches to answer questions. You can answer a question with many turns of search and reasoning.

Based on the history information, you need to suggest the next action to complete the task. 
You will be provided with:
1. Your history search attempts: query in format <search> query </search> and the returned search results in <information> and </information>.
2. The question to answer.

IMPORTANT: You must strictly adhere to the following rules:
1. Choose ONLY ONE action from the list below for each response, DO NOT perform more than one action per step.
2. Follow the exact syntax format for the selected action, DO NOT create or use any actions other than those listed.
3. **Don't do duplicate search.** Pay attention to the history search results.

Valid actions:
1. <search> query </search>: search the web for information if you consider you lack some knowledge.
2. <answer> answer </answer>: output the final answer if you consider you are able to answer the question. The answer should be short and concise. No justification is needed.
3. <summary> important parts of the history turns </summary>: summarize the history turns. Reflect the search queries and search results in you history turns, and keep the information you consider important for answering the question and generating your report. Still keep the tag structure, keep search queries between <search> and </search>, and keep search results between <information> and </information>. The history turn information for your subsequent turns will be updated accoring to this summary action.

Format:
You should pay attention to the format of your output. You can choose **ONLY ONE** of the following actions:
    - If You want to search, You should put the query between <search> and </search>. 
    - If You want to summarize the history turns, You should put the summary between <summary> and </summary>.
    - If You want to give the final answer, You should put the answer between <answer> and </answer>.
    You can only use ONE action per response.

Note: text between <information></information> is the search results from search engine after you perform a search action, **DO NOT** include any information in <information></information> in your output.

Question: {question}

History Turns: (empty if this is the first turn)

\end{code}
\end{tcolorbox}

\begin{tcolorbox}[title=Agent Framework Prompt for Models without Internal Thinking,
                  colback=gray!5, colframe=gray!50,
                  breakable, enhanced]
\begin{code}
Your are a research assistant with the ability to perform web searches to answer questions. You can answer a question with many turns of search and reasoning.

Based on the history information, you need to suggest the next action to complete the task. 
You will be provided with:
1. Your history search attempts: query in format <search> query </search> and the returned search results in <information> and </information>.
2. The question to answer.

IMPORTANT: You must strictly adhere to the following rules:
1. Choose ONLY ONE action from the list below for each response, DO NOT perform more than one action per step.
2. Follow the exact syntax format for the selected action, DO NOT create or use any actions other than those listed.
3. **Don't do duplicate search.** Pay attention to the history search results.

Valid actions:
1. <search> query </search>: search the web for information if you consider you lack some knowledge.
2. <answer> answer </answer>: output the final answer if you consider you are able to answer the question. The answer should be short and concise. No justification is needed.
3. <summary> important parts of the history turns </summary>: summarize the history turns. Reflect the search queries and search results in you history turns, and keep the information you consider important for answering the question and generating your report. Still keep the tag structure, keep search queries between <search> and </search>, and keep search results between <information> and </information>. The history turn information for your subsequent turns will be updated accoring to this summary action.

Format:
You should pay attention to the format of your output. You can choose **ONLY ONE** of the following actions:
    - If You want to search, You should put the query between <search> and </search>. 
    - If You want to summarize the history turns, You should put the summary between <summary> and </summary>.
    - If You want to give the final answer, You should put the answer between <answer> and </answer>.
    You can only use ONE action per response.

Format:
<think> thinking process </think>
[your action output]

Example:
<think> I need to answer the question, so I need to... </think>
<search> query </search>

Note: text between <information></information> is the search results from search engine after you perform a search action, **DO NOT** include any information in <information></information> in your search action.

Question: {question}

Question: {question}

History Turns: (empty if this is the first turn)

\end{code}
\end{tcolorbox}

\subsection{Prompt For Behavior Identification}
\label{sec: rule_prompt}

\begin{tcolorbox}[title=Prompt for Trajectory Analysis,
                  colback=gray!5, colframe=gray!50,
                  breakable, enhanced]
\begin{code}
[Instruction]
You are tasked with analyzing multi-step trajectories of a search agent's two attempts for answering the same question using search tools. One of the attempts correctly answers the question, and another attempt does not. Based on the content, please provide a detailed explanation of why one attempt succeeds and the other fails.
There are two parts in each step of the trajectory:
1. Agent output: The agent's output in this step, consists of it's thinking process and the final action.
2. Environment feedback: The feedback from the environment, including the search results wrapped in <information> and </information> tags when the agent performs a search action in this step.

The agent could perform one of the following actions in each step:
1. <search> query </search>: search the web for information
2. <answer> answer </answer>: output the final answer
3. <summary> important parts of the history turns </summary>: summarize the history turns to keep valuable information for solving the question.

Please analyze the agent's behavior in each step and provide a detailed explanation of why one attempt succeeds and the other fails.

[Question]
{question}

[Trajectory 1]
{trajectory_1}

[Evaluation Results 1]
{evaluation_results_1}

[Trajectory 2]
{trajectory_2}

[Evaluation Results 2]
{evaluation_results_2}

[Your Explanation]
\end{code}
\end{tcolorbox}

\begin{tcolorbox}[title=Prompt for Key Reasoning Behavior Extraction,
                  colback=gray!5, colframe=gray!50,
                  breakable, enhanced]
\begin{code}
You are an expert in analyzing the behavior of a search agent. You will be provided with an explanation about a search agent's two attempts to answer the same question using search tools. The first attempt correctly answers the question, while the second attempt fails.  

Based on the explanation of why trajectory 1 succeeds while trajectory 2 fails, extract the key reasoning behaviors statements implied by the explanation that lead to the success of trajectory 1. These should be clear, objective, and unambiguously verifiable.  

Return the list as a JSON array of strings. Do not include markdown code fences. If there are no rule-like statements, return an empty JSON array.

[Reasoning]  
{reasoning_text}
\end{code}
\end{tcolorbox}

\begin{tcolorbox}[title=Prompt for Behavior Summarization,
                  colback=gray!5, colframe=gray!50,
                  breakable, enhanced]
\begin{code}
You are an expert in analyzing the behavior of a search agent. You are provided with a set of behaviors describing the the reasoning process and actions of the agent.  

Below is a list of behaviors regarding the behavior of the search agent. Some behaviors may be duplicates or express very similar meanings. Please merge them by removing duplicates and consolidating similar behaviors, while keeping only the most essential information. When merging, discard narrow or overly specific restrictions, and retain only general behaviors that are broadly applicable.  

The final rules should be clear, objective, and unambiguous, so they can be reliably used to evaluate the agent's reasoning and interaction trajectory.  

Return the merged list as a JSON array of strings. Do not include markdown code fences.  

[Behaviors]  
{behaviors_text}
\end{code}
\end{tcolorbox}

\subsection{Prompt For Behavior Frequency Evaluation}
\label{sec: behavior_prompt}

\begin{tcolorbox}[title=Prompt For Behavior Frequency Evaluation,
                  colback=gray!5, colframe=gray!50,
                  breakable, enhanced]
\begin{code}
[Instruction]
You are tasked with analyzing a multi-step trajectory of a search agent's attempt for answering a question using search tools. 

The agent can perform one of the following actions in each step:
1. <search> query </search>: search the web for information
2. <answer> answer </answer>: output the final answer
3. <summary> important parts of the history turns </summary>: summarize the history turns to keep valuable information for solving the question.

There are two parts in each step of the trajectory:
1. Agent output: The agent's output in this step, consists of it's thinking process and the final action.
2. Environment feedback: The feedback from the environment, including the search results wrapped in <information> and </information> tags when the agent performs a search action in this step.

Please act as an judge to evaluate whether the agent's thinking process and actions in this trajectory demonstrated any of following behaviors:

**behavior1: Information Verification**
The agent validates information across multiple reliable sources to ensure its conclusions are well-founded.
* **Cross-Referencing:** Actively seeking out and comparing multiple sources to confirm critical facts, or performing additional searches to verify the information.
* **Citing Evidence:** Explicitly basing its reasoning and conclusions on the information found, rather than making unsupported claims.

**behavior2: Authority Evaluation**
The agent assesses the reliability of its sources and resolves conflicting information.
* **Detecting Conflicts:** Identifying when different sources provide conflicting information and attempting to resolve the discrepancy.
* **Prioritizing Authority:** Giving more weight to official documentation, academic papers, and reputable news outlets over forums, blogs, or less reliable sources.

**behavior3: Adaptive Search**
The agent intelligently modifies its search strategy based on the information and challenges encountered in previous steps.
* **Narrowing Focus:** Using initial broad search results to identify more specific and effective keywords for subsequent searches.
* **Broadening Scope:** Widening the search terms or approach when initial queries are too narrow and yield no useful results.

**behavior4: Error Recovery**
The agent recognizes previous errors and takes actions to correct its course.
* **Acknowledging Failure:** Explicitly noting when a search query or an entire strategy is not yielding useful information, or some mistakes are made.
* **Strategic Pivoting:** Decisively abandoning a failed approach and formulating a new plan to achieve the user's goal, or taking actions to correct the mistakes.

Be as objective as possible when evaluating the behaviors and do not evaluate other characteristics of the response. If the behavior is not applicable for this task, treat it as if the behavior is not demonstrated.

You must provide your answer with the following json format without markdown code fences:

{
  "behavior1": "<'Yes' or 'No'>",
  "behavior2": "<'Yes' or 'No'>",
  "behavior3": "<'Yes' or 'No'>",
  "behavior4": "<'Yes' or 'No'>",
  ...
}

[Question]
{question}

[Trajectory]
{trajectory}

[Your Answer]
\end{code}
\end{tcolorbox}

\subsection{Prompt for Evaluation}
\label{sec: prompt_for_eval}

\begin{tcolorbox}[title=Prompt For LLM-Judge Evaluation on Web Tasks,
                  colback=gray!5, colframe=gray!50,
                  breakable, enhanced]
\begin{code}
    
Please determine if the predicted answer is SEMANTICALLY equivalent to the labeled answer. 
Question:  {question} 
Labeled Answer:  {gt_answer} 
Predicted Answer: {pred_answer}  

{
"rationale": "your rationale for the judgement, as a text", 
"judgement": "your judgement result, can only be 'correct' or 'incorrect'" 
}
\end{code}
\end{tcolorbox}

\begin{tcolorbox}[title=Prompt For LLM-Judge Evaluation on Multi-hop QA Tasks,
                  colback=gray!5, colframe=gray!50,
                  breakable, enhanced]
\begin{code}
    
You will be given a question and its ground truth answer list where each item can be a ground truth answer. Provided a pred_answer, you need to judge if the pred_answer correctly answers the question based on the ground truth answer list.
You should first give your rationale for the judgement, and then give your judgement result (i.e., correct or incorrect).

Here is the criteria for the judgement:
1. The pred_answer doesn't need to be exactly the same as any of the ground truth answers, but should be semantically same for the question.
2. Each item in the ground truth answer list can be viewed as a ground truth answer for the question, and the pred_answer should be semantically same to at least one of them.

question: {question}
ground truth answers: {gt_answer}
pred_answer: {pred_answer}

The output should in the following json format:
{ 
"rationale": "your rationale for the judgement, as a text", 
"judgement": "your judgement result, can only be 'correct' or 'incorrect' 
}
\end{code}
\end{tcolorbox}

\section{Case Study}

Here is a case study for how the behavior-primed model (Qwen3 1.7B + Behavior Prime) successfully solves a question, but the non behavior-primed model fails (Qwen3 1.7B + Direct RL). The question is from the GAIA dataset. The trajectory for both models is provided below.

\paragraph{Question} 
``What was the actual enrollment count of the clinical trial on H.~pylori in acne vulgaris patients from Jan--May 2018 as listed on the NIH website?''

\subsection{Successful Trajectory}

\begin{tcolorbox}[title=Qwen3 1.7B + Behavior Prime,
colback=gray!5, colframe=gray!50,
breakable, enhanced]
\textbf{Step 1:} \\
\high{<think>}
\begin{code}
Okay, so the user's looking for the enrollment count of a specific clinical trial on 
H. pylori and acne vulgaris, but they're zeroing in on the period from January to May of 2018, 
and they're specifically mentioning the NIH website. ...
\end{code}

\highlightA{
\textit{[Behavior 1: Information Verification]} \\
My task is clear: verify if the quoted text exactly matches Greetham's article. 
Accuracy is paramount; I'll use 'uncoupled', 'authors', 'mis-transmission', and 'veil' 
to zero in on the relevant section.
}

\high{</think>}\\

\high{<search>}
\begin{code}
clinical trial H. pylori acne vulgaris enrollment Jan-May 2018 NIH website
\end{code}
\high{<search>}
\begin{code}
<information> retrieved documents </information>
\end{code}

\textbf{Step 2:} \\
\high{<think>}
\begin{code}
I've got a few hits, though. Result #2 ... looks promising because it's linked to ClinicalTrials.gov, 
which is where I'd expect to find this kind of information. ...
\end{code}

\highlightB{
\textit{[Behavior 2: Authority Evaluation]} \\
I’m aiming for the USGS’s own reports or databases, like the ‘Nonindigenous Aquatic Species’ 
page, to get the most reliable data.}
\high{</think>}

\high{<search>} 
\begin{code}
NCT03411733 enrollment count Jan-May 2018 NIH 
\end{code}
\high{</search>}
\begin{code}
<information> retrieved documents </information>
\end{code}

\textbf{Step 3:} \\
\high{<think>}
\begin{code}
The problem is that the search engine is not returning the number directly. 
It's likely that the information is not readily available in the snippets. 
So, I need to try to get to the page for NCT03411733 ...
\end{code}
\highlightC{
\textit{[Behavior 3: Adaptive Search]} \\
The search engine may not have indexed the quote perfectly, or the user’s quote may differ slightly. 
I’ll refine my strategy.} \\
\high{</think>}

\high{<search>} 
\begin{code}
NCT03411733 trial start date and recruitment numbers Jan-May 2018 NIH
\end{code}
\high{</search>}
\begin{code}
<information> retrieved documents </information>
\end{code}

\textbf{Step 4:} \\
\high{<think>}
\begin{code}
The last query didn’t bring me directly to the ClinicalTrials.gov record. 
I need to correct my approach and focus on the trial ID itself.
\end{code}

\highlightD{
\textit{[Behavior 4: Error Recovery]} \\
My initial search didn’t deliver the goods... I need to get more precise.} \\
\high{</think>}

\high{<search>} \\
NCT03411733 actual enrollment count Jan-May 2018 NIH \\
\high{</search>}\\
\begin{code}
<information> retrieved documents </information>
\end{code}

\textbf{Step 5:} \\

\high{<think>}
\begin{code}
...
\end{code}
\high{</think>} \\
\high{<answer>}
\begin{code}
The actual enrollment count is 90 patients. 
\end{code}
\high{</answer>} \\
\textbf{Result: Correct.} \\
\end{tcolorbox}

\subsection{Unsuccessful Trajectory}
\begin{tcolorbox}[title=Qwen3 1.7B + Direct RL,
colback=gray!5, colframe=gray!50, breakable, enhanced]
\textbf{Turn 1} \\
\high{<think>}
\begin{code}
Okay, let's tackle this question. The user is asking for the actual enrollment count of a clinical trial on 
H. pylori in acne vulgaris patients from January to May 2018, specifically as listed on the NIH website.

Since the history is empty, I should start by searching for the clinical trial. 
The valid action here is to perform a search. 
The query would be something like "H. pylori clinical trial acne vulgaris enrollment Jan-May 2018 NIH".

I'll use the <search> action with that query. The search results might show the trial's details, including the enrollment numbers.
I need to make sure the NIH website has this information.
\end{code}
\high{</think>}

\high{<search>}
\begin{code}
H. pylori clinical trial acne vulgaris enrollment Jan-May 2018 NIH
\end{code}
\high{</search>}

\begin{code}
<information> ... Found NCT03411733 and related links ... </information>
\end{code}

\infohighlight{No attempt to cross-check, only one generic search.}

\textbf{Turn 2} \\
\high{<think>}
\begin{code}
[No reasoning; model attempts an invalid summary action.]
\end{code}
\high{</think>}

\high{<summary>}
\begin{code}
<search> H. pylori clinical trial acne vulgaris enrollment Jan-May 2018 NIH </search>
<information>[All 10 results copied verbatim]</information>
\end{code}
\high{</summary>}

\infohighlight{Model generated an invalid action format. 
Instead of correcting, it collapsed history incorrectly.}

\textbf{Turn 3} \\
\high{<think>}
\begin{code}
[No new reasoning. Model repeats identical search query from Turn 1.]
\end{code}
\high{</think>}

\high{<search>}
\begin{code}
H. pylori clinical trial acne vulgaris enrollment Jan-May 2018 NIH
\end{code}
\high{</search>}

\begin{code}
<information> ... Same results again ... </information>
\end{code}

\infohighlight{Query was repeated with no refinement.}

\textbf{Turn 4} \\
\high{<think>}
\begin{code}
Prematurely concludes the answer without verifying enrollment numbers.
\end{code}
\high{</think>}

\high{<answer>}
\begin{code}
The NIH clinical trial on H. pylori and acne vulgaris from January to May 2018 (NCT03411733) 
had an enrollment count of 100 participants. 
\end{code}
\high{</answer>}
\newpage
\textbf{Result: Incorrect. Extracted "100" without checking against 
the authoritative NIH record. NIH/ClinicalTrials.gov was named, but not validated.} \\
\end{tcolorbox}






\end{document}